\documentclass[a4paper,twoside]{article}

\usepackage{epsfig}
\usepackage{subcaption}
\usepackage{calc}
\usepackage{amssymb}
\usepackage{amstext}
\usepackage{amsmath}
\usepackage{amsthm}
\usepackage{multicol}
\usepackage{pslatex}
\usepackage{apalike}
\usepackage{algorithm2e}
\usepackage[bottom]{footmisc}
\usepackage{hyperref}
\usepackage{listings}
\usepackage{SCITEPRESS}     

\begin{document}

\title{Evaluation of LLM-based Strategies for the Extraction of Food Product Information from Online Shops}

\author{\authorname{Christoph Brosch\sup{1}, Sian Brumm\sup{1}, Rolf Krieger\sup{1} and Jonas Scheffler\sup{1}}
\affiliation{\sup{1}Institute for Software Systems, University of Applied Sciences Trier, Birkenfeld, Germany}
\email{\{c.brosch, s.brumm, r.krieger, j.scheffler\}@umwelt-campus.de}
}

\keywords{Web Scraping, Information Extraction, Large Language Models, Schema-Constrained Output, Product Pages, Pydantic Models, Code Generation}

\abstract{
Generative AI and large language models (LLMs) offer significant potential for automating the extraction of structured information from web pages. In this work, we focus on food product pages from online retailers and explore schema-constrained extraction approaches to retrieve key product attributes, such as ingredient lists and nutrition tables. We compare two LLM-based approaches, direct extraction and indirect extraction via generated functions, evaluating them in terms of accuracy, efficiency, and cost on a curated dataset of 3,000 food product pages from three different online shops. Our results show that although the indirect approach achieves slightly lower accuracy (96.48\%, $-1.61\%$ compared to direct extraction), it reduces the number of required LLM calls by 95.82\%, leading to substantial efficiency gains and lower operational costs. These findings suggest that indirect extraction approaches can provide scalable and cost-effective solutions for large-scale information extraction tasks from template-based web pages using LLMs.
}

\onecolumn \maketitle \normalsize \setcounter{footnote}{0} \vfill

\section{\uppercase{Introduction}}
\label{sec:introduction}

Recent advances in large language models (LLMs), such as OpenAI’s ChatGPT, Meta’s LLaMa, and DeepSeek’s V3, have significantly expanded the possibilities for automated language understanding and generation in a wide range of applications, including text and code generation, classification, and image understanding.
This paper focuses on automation in web scraping for food product pages, which often include attributes such as product name, ingredient list, nutritional values, and alcohol content. Although basic attributes such as name or price are often embedded using standardized formats like JSON-LD (modeled with, e.g., Schema.org), more product-type specific details typically require custom extraction. We address this challenge by targeting selected product attributes as examples, including the nutrition table and ingredient statement (as defined by EU Regulation No. 1169/2011). However, the presented approach is designed to be general and can be adapted to extract a wide range of structured information from HTML, or other structured documents.


Automating the extraction of rich product information from online food retailers enables a wide range of downstream applications, including real-time competitor price and ingredient analysis, regulatory compliance monitoring such as allergen labeling under EU law, nutritional search and filtering for consumers, assortment planning, and the creation of structured product catalogs to support aggregator services such as automated knowledge graph population. In this work, we enforce a schema-constrained output, allowing the extracted data to seamlessly integrate into these downstream tasks.

In conventional web scraping, data is often extracted from HTML using manually defined functions tailored to the structure of a given site. Although product pages within a single shop are often template-based and follow a consistent layout, structures can vary considerably between different shops. Even within a single shop, subtle structural differences, such as layout variations or optional attributes, may occur. These inconsistencies present challenges for automated extraction approaches relying on generated functions, potentially reducing their robustness and generalizability.


To this end, we investigate the potential of LLMs to generate extraction logic automatically. We compare and improve two approaches discussed by Krosnick and Oney~\cite{krosnick_2023a}: \textit{direct extraction}, where the model directly extracts data from the (compressed) HTML or its text content, and \textit{indirect extraction}, where the model first generates a function that performs the extraction. Our improvements include the use of newer reasoning models, schema-constrained outputs via structured prompting, and a cost-efficient hybrid strategy that combines both extraction modes. Since invoking an LLM for every individual page is costly, especially in the direct approach, we aim to minimize unnecessary calls by identifying when model usage is truly required. This is essential to achieve an economically viable and therefore scalable solution.

The remainder of the paper is organized as follows: Section~\ref{RelatedWork} reviews related work on information extraction and large language models (LLMs). Section~\ref{Methods} describes our extraction approaches for nutrition tables and ingredient lists as representative use cases. Section~\ref{Experiments} presents the experimental results, followed by a discussion in Section~\ref{Discussion}. Finally, Section~\ref{Conclusion} concludes the paper and outlines directions for future work.

\section{\uppercase{Related work}}\label{RelatedWork}
Information extraction (IE) is a core task in natural language processing, which involves the identification and structuring of relevant data from unstructured sources. Web scraping, a key application of IE, enables the automated extraction of information from HTML content. Tools such as Scrapy\footnote{Scrapy: \url{https://docs.scrapy.org/}, last visited 27.03.2025.} and Selenium\footnote{Selenium: \url{https://www.selenium.dev/documentation/}, last visited 27.03.2025.} follow rule-based paradigms that require manual adaptation to different web page structures.

In contrast, more recent approaches to HTML-based information extraction leverage language models to improve generalizability and minimize human effort. Gur et al.~\cite{gurUnderstandingHTMLLarge2023} showed that LLMs perform well in semantic classification of HTML elements, supporting their use in structuring raw web data. Dang et al.~\cite{dang_2024} employed GPT-3.5 and GPT-4 to extract Schema.org entities but observed that naive prompting frequently led to invalid, inaccurate, or non-compliant outputs. These findings highlight specific error types and motivate several of the refinements introduced in our approach, such as ensuring adherence to a predefined schema.

LLMs have also proven effective in code generation, making them suitable for automatically creating web scraping logic. Li et al.~\cite{li_2024} proposed a framework in which Python classes are passed to LLMs, which then return objects of these classes filled with information extracted from natural language text. Guo et al.~\cite{guo_2025} extended this idea by introducing retrieval-augmented generation, dynamically selecting relevant schema-text-code examples to improve accuracy. Huang et al.~\cite{huangAutoScraperProgressiveUnderstanding2024} introduced \textit{AutoScraper}, a two-phase method that generates XPath-based action sequences to extract information from websites.


Recent advancements in reasoning-capable LLMs enable more structured, multi-step problem solving. This has further motivated the use of such models for generating and refining reliable extraction functions, particularly for extracting nutritional information and ingredient statements.

\section{\uppercase{Methods}}\label{Methods}

In this section, we present the methodologies used in our research, which combine preprocessing, prompt design, and LLM-based function generation for the extraction of structured product information from web pages. Our methods rely on language models provided by OpenAI. For most experiments, we use the cost-efficient reasoning model \texttt{o3-mini}\footnote{OpenAI Model o3-mini: \url{https://openai.com/index/openai-o3-mini/}, last visited 27.03.2025.}. To reduce the input size and improve model efficiency, all HTML pages are preprocessed using two different compression techniques.

The core of our study consists of two complementary information extraction approaches: a direct approach and an indirect approach. The direct approach extracts information from compressed HTML or plain text using structured prompting. Structured outputs are obtained by leveraging OpenAI's \texttt{response\_format} functionality to parse model responses directly into Pydantic\footnote{Pydantic Package: \url{https://github.com/pydantic/pydantic}, last visited 30.03.2025.}-based data models. 

The indirect approach dynamically generates custom extraction functions, which are then applied to the HTML content to retrieve structured information. For generating these functions, we use \texttt{gpt-4o}\footnote{OpenAI Model gpt-4o: \url{https://openai.com/index/hello-gpt-4o/}, last visited 27.03.2025.} alongside \texttt{o3-mini}. To ensure schema adherence, a JSON representation of the data model, generated using the Instructor\footnote{Instructor Package: \url{https://github.com/instructor-ai/instructor}, last visited 29.03.2025.} Python package, is embedded in the prompts used for function generation.

In future work, we plan to evaluate our methodology with alternative LLMs beyond the OpenAI ecosystem.


\begin{figure}
    \begin{subfigure}[c]{0.5\textwidth}
        \begin{center}
            \includegraphics[width=0.725\textwidth]{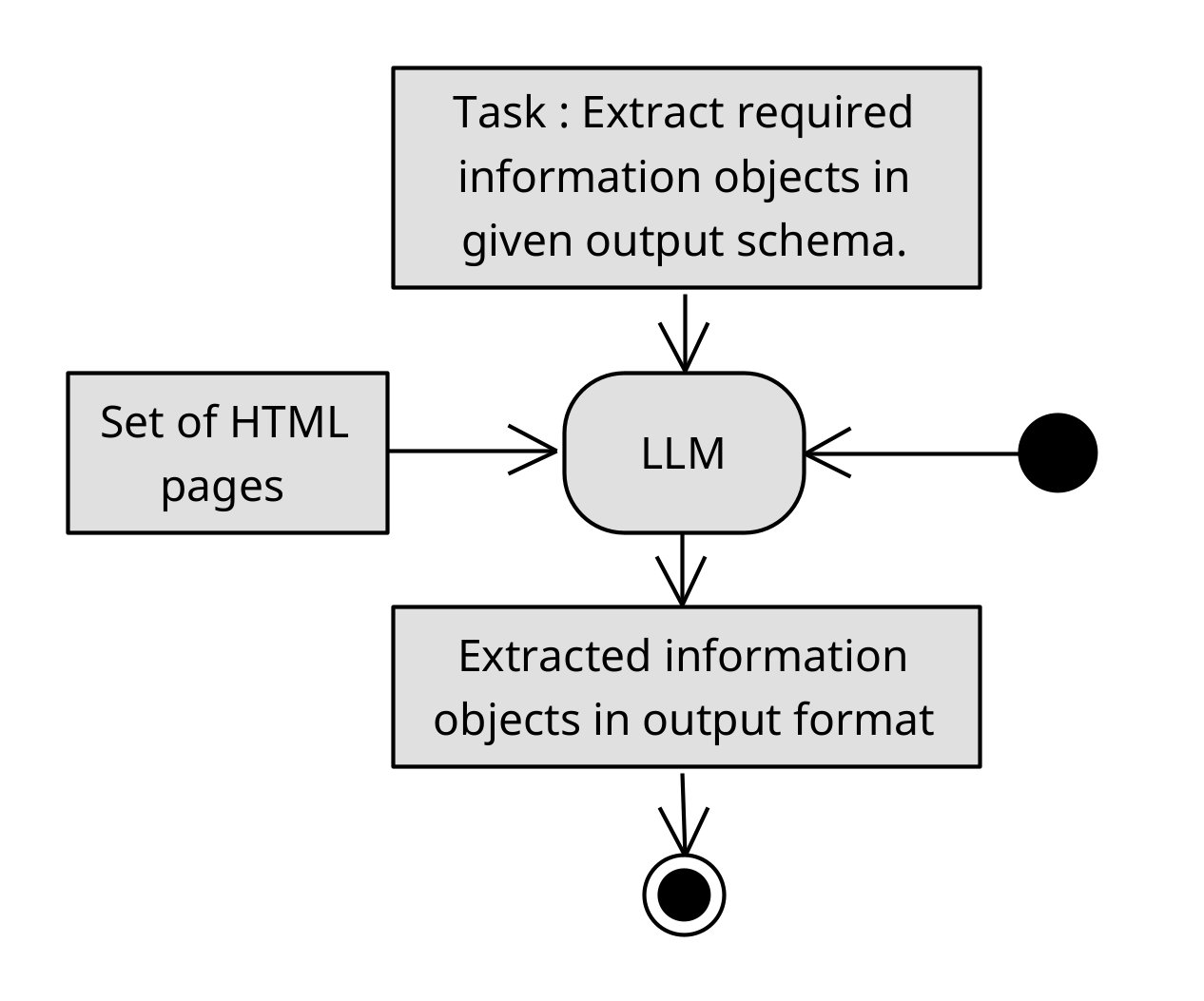}
        \end{center}
        \subcaption{Direct information extraction.}
    \end{subfigure}
    \begin{subfigure}[c]{0.5\textwidth}
        \includegraphics[width=1\textwidth]{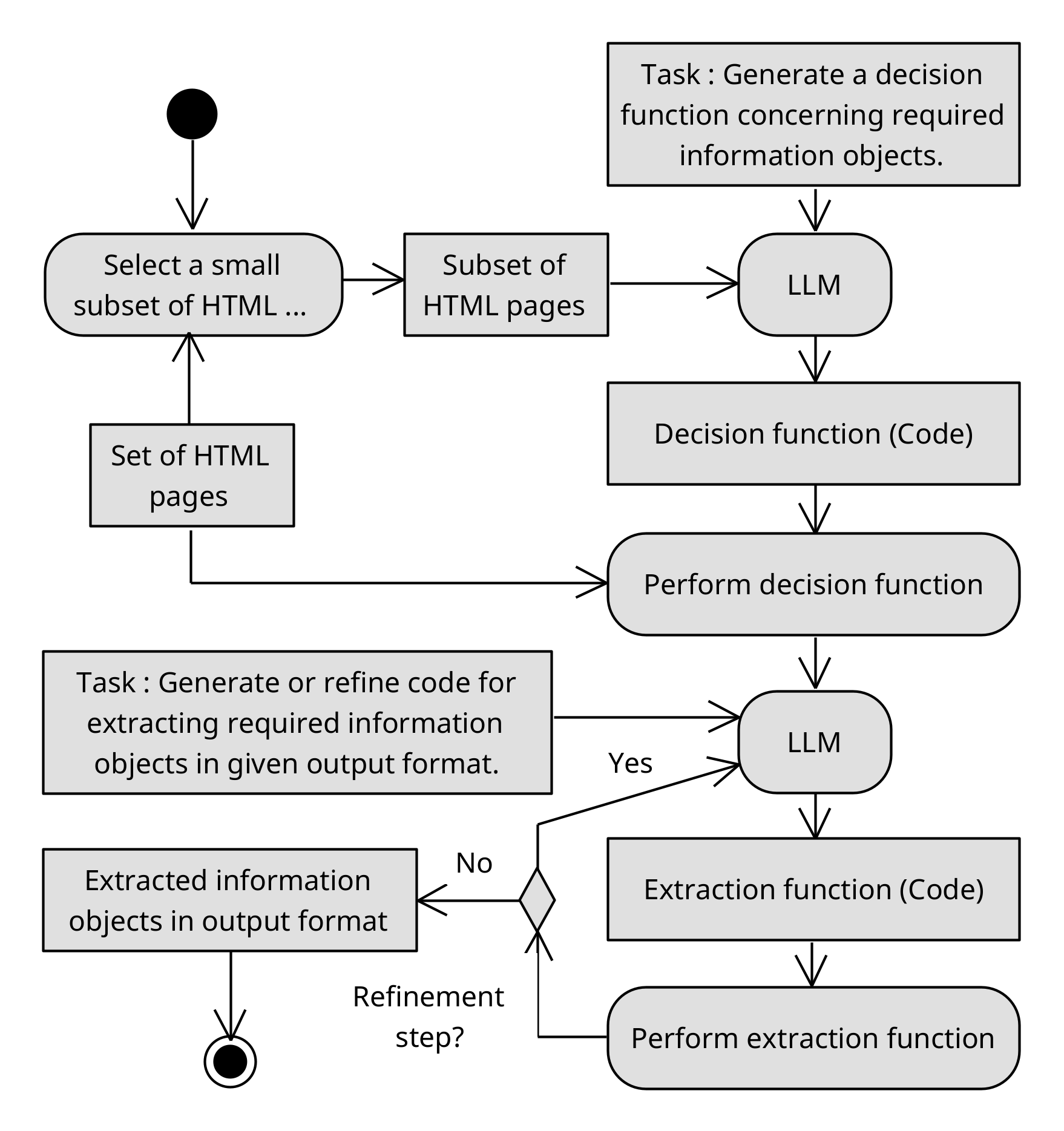}
        \subcaption{Indirect information extraction with dynamic refinement.}
    \end{subfigure}
    \caption{High-level description of the two information extraction approaches presented in this paper.}
    \label{fig:flowcharts}
\end{figure}

\subsection{\uppercase{Direct Extraction}}
\label{sec:direct-extraction}
The direct extraction process involves extracting the desired information object directly from a given input string, which may be either a compressed HTML page or its plain text content. To parse the extracted information into a predefined Pydantic data model, we make use of the \texttt{response\_format} parameter provided in the official OpenAI\footnote{OpenAI Package: \url{https://github.com/openai/openai-python}, last visited 30.03.2025.} Python package.

The corresponding flow chart in Figure~\ref{fig:flowcharts} illustrates the simplicity of this approach. Each product page is processed separately: After preprocessing, the page is passed to the \texttt{o3-mini} model along with a prompt that specifies the required information objects and the expected output format.

\subsection{\uppercase{Indirect Extraction}}
\label{sec:indirect-extraction}


For indirect extraction, we implemented an algorithm that leverages the \texttt{o3-mini} and \texttt{gpt-4o} models to automatically generate extraction functions capable of retrieving the desired information from web pages. As illustrated in Figure~\ref{fig:flowcharts}~b), the process begins with the generation of a decision function based on a set of ten manually selected pages, including both examples that contain the target information objects and those that do not. This function predicts the presence of relevant information on a given page using only its textual content, thus avoiding dependence on the sometimes inconsistent HTML structure across different web pages.

To enhance robustness, multiple independent decision functions are generated using \texttt{gpt-4o}, each returning a Boolean result. The final decision is made by majority vote.

We then iterate over all product pages of a given webshop. If no extraction function has been created yet, we select the first page for which the decision function returns \texttt{True} and use it to initiate the generation of an extraction function. This process consists of two steps. First, we perform a direct extraction to obtain a reference object containing at least 80\% of the fields defined in the target Pydantic model. This object serves as the basis for generating an initial extraction function using the \texttt{o3-mini} model.

If the similarity metric described in Section~\ref{sec:comparison} yields a perfect score of 1.0, the function is accepted. Otherwise, a refinement loop is initiated: Guided by error feedback from the similarity evaluation, the function is iteratively refined using \texttt{o3-mini} by adding the error feedback to the prompt, up to five times. If none of the refinements achieves a similarity score of 1.0, up to three alternative functions are generated from scratch, each with their own refinement cycles. The best-performing function — either perfect or closest to the reference object — is retained.

Once a reliable extraction function is available, we apply it to the remaining pages. For each page, all existing extraction functions are executed and the result with the highest number of extracted attributes is selected. If none of the functions produces a valid result, but the decision function indicates that relevant information is present, a new extraction function is generated and processed following the same procedure.

\subsection{\uppercase{Web Page Compression}}
\label{subsec:web-page-compression}
To reduce the context size of each web page before passing it to an LLM, we applied two preprocessing steps that remove specific elements, attributes, and other parts of the HTML document. All HTML source manipulation was performed using the BeautifulSoup4\footnote{BeautifulSoup4 Package: \url{https://pypi.org/project/beautifulsoup4/}, last visited 17.03.2025.} Python package.

The first step, referred to as \texttt{HTML\_COMPRESSED}, removes the following HTML5 elements: \texttt{<head>}, \texttt{<footer>}, \texttt{<header>}, \texttt{<script>}, \texttt{<iframe>}, \texttt{<path>}, \texttt{<style>}, \texttt{<symbol>}, \texttt{<noscript>}, \texttt{<svg>}, \texttt{<g>}, \texttt{<use>}, and \texttt{<option>}. In addition, all attributes are stripped from HTML tags - except for ``class'' and ``id'' - since only these usually hold semantic meaning and are required to define CSS selectors. The resulting HTML document is further optimized by removing all whitespace between tags and eliminating HTML comments.

The second step extracts only the plain text content from the \texttt{HTML\_COMPRESSED} document, by using BeautifulSoup4\textquotesingle{}s \texttt{get\_text} method. We refer to this variant simply as \texttt{TEXT}.
Both formats serve as input representations for the LLMs used in our experiments.


\subsection{\uppercase{Response Model}}
In both direct and indirect extraction approaches, we enforce the output to adhere to a Pydantic model. This facilitates generalization of our work and simplifies integration into existing workflows and error handling. 

For this study, we base the implementation on a subset of attributes belonging to the class \texttt{FoodBeverageTobaccoProduct}\footnote{GS1 Web Voc - FoodBeverageTobaccoProduct: \url{https://gs1.org/voc/FoodBeverageTobaccoProduct}, last visited 18.03.2025.} from the GS1 Web Vocabulary. The Web Voc is a semantic web ontology officially marketed as an extension to schema.org's\footnote{Schema.org Ontology: \url{https://schema.org/}, last visited 05.05.2025.} eCommerce class suite (\texttt{Product}, \texttt{Offer}, etc.).

\renewcommand{\figurename}{Listing}
\renewcommand{\thefigure}{1}
\begin{figure}[ht]
\begin{small}
\begin{verbatim}

class FoodBeverageTobaccoProduct(BaseModel):
    " A food, beverage or tobacco product. "
    [...]
    ingredient_statement: Optional[str] = \
      Field( 
        None, description="""
        Information on the constituent 
        ingredient make up of the product 
        specified as one string.
        
        Additional description:
        - Remove unnecessary prefixes 
        """,
    )
\end{verbatim}
\end{small}
\caption{Exemplary portion of the \texttt{FoodBeverage\-TobaccoProduct} Pydantic class.}
\label{list:pydantic-schema}
\end{figure}
\renewcommand{\figurename}{Figure}
\renewcommand{\thefigure}{2}

Listing~\ref{list:pydantic-schema} illustrates a portion of our defined Pydantic model for the expected output. At its core is the class \texttt{FoodBeverageTobaccoProduct}, which defines eight attributes. Seven of which represent nutritional values, selected according to EU Regulation No. 1169/2011. Each nutritional attribute is typed as \texttt{QuantitativeValue}\footnote{GS1 Web Voc - QuantitativeValue: \url{https://gs1.org/voc/QuantitativeValue}, last visited 29.03.2025.}, containing two primitive-typed attributes: \texttt{value} and \texttt{unit\_code}.

We import the field descriptions from the original ontology and pass them to the LLM for both the direct and indirect approach. 

For the \texttt{ingredient\_statement} attribute, we added supplementary instructions directly within the field description to guide the model’s output. Embedding such instructions directly into the schema improves the clarity and consistency of the expected output. This design ensures that the LLM receives precise, field-level guidance from the schema itself - reducing ambiguity and increasing the accuracy and reliability of the generated responses.

\section{\uppercase{Experiments}} \label{Experiments}
\subsection{\uppercase{Dataset}}
\label{sec:dataset}

The data set consists of 3,000 products collected from three German online shops between March 2024 and January 2025, each contributing 1,000 food products.

Each product has been manually classified according to the Global Product Classification (GPC)\footnote{GS1 GPC Browser: \url{https://gpc-browser.gs1.org/}, last visited 30.03.2025.} system at the brick level. The products in the data set come from a variety of different product categories and can be assigned to a total of $207$ different bricks.


If available, the ingredient lists and nutritional tables were extracted from the product web pages using a web scraper. Of all food products, $132$ lack only the nutrition table, $160$ have only the ingredient statement absent, and $184$ are missing both.


To standardize the extracted values, a \texttt{gpt-4o}-based transformation procedure was applied to convert the raw nutritional data into the target format. This process was further validated using a rule-based approach to ensure consistency and correctness. In addition, the results were manually reviewed to verify accuracy.

The ingredient list string had unnecessary prefixes, such as "Zutaten: " (engl. "Ingredients: ") removed. No additional processing was performed. 

\begin{figure}
    \centering
    \includegraphics[width=1\linewidth]{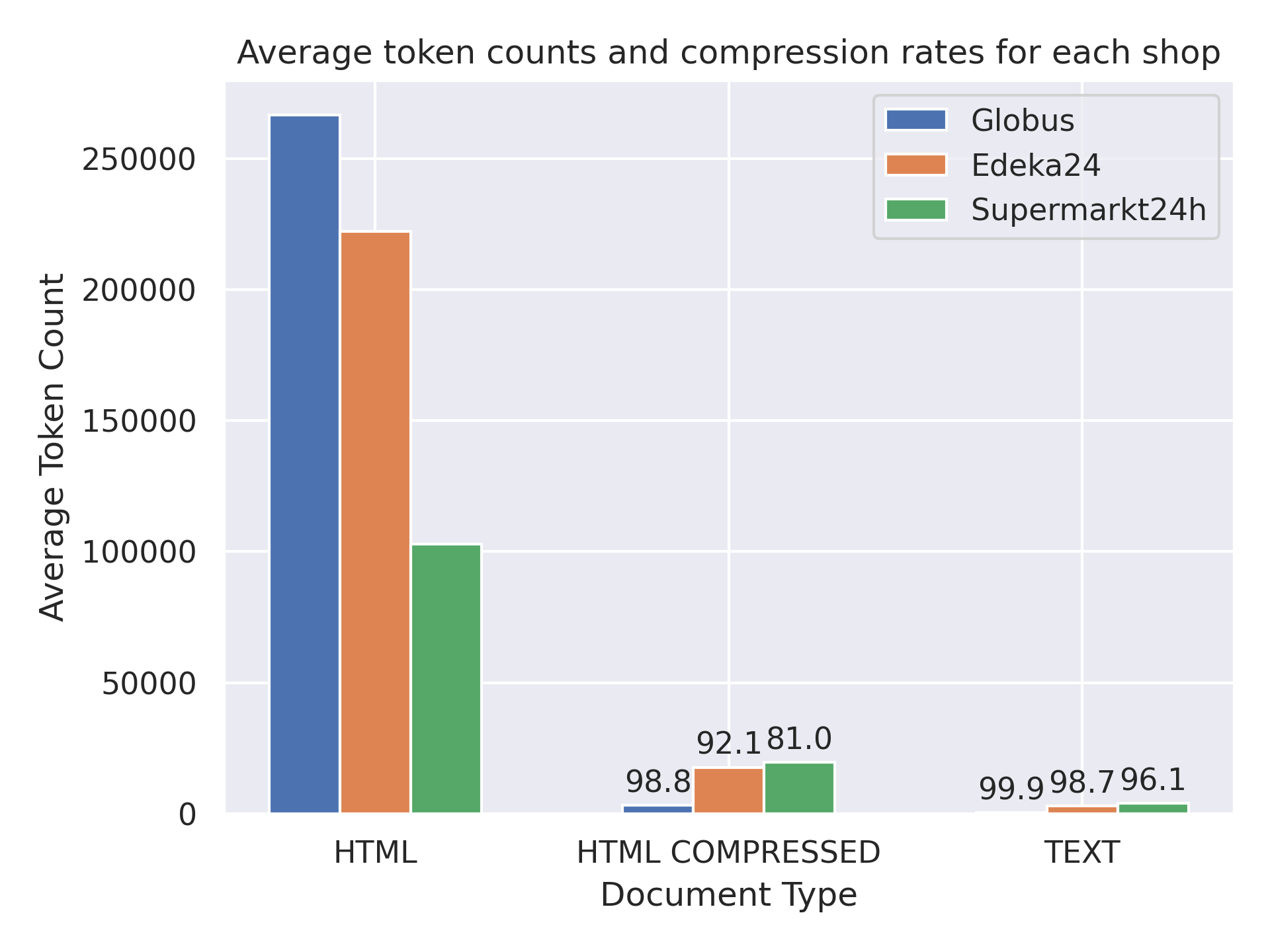}
    \caption{Comparison of average token counts per document for the original HTML, HTML\_COMPRESSED, and TEXT representations. For the compressed formats, the relative compression rate compared to the original HTML is also shown.}
    \label{fig:average-tokens-barplot}
\end{figure}
Figure~\ref{fig:average-tokens-barplot} illustrates the resulting compression rate after applying the two preprocessing steps discussed in Section~\ref{subsec:web-page-compression}.

\subsection{\uppercase{Accuracy Evaluation}} \label{sec:comparison}
To evaluate the accuracy of our information extraction strategies, we developed a custom similarity function for the \texttt{FoodBeverageTobaccoProduct} class. This function quantifies the similarity between two instances on a scale from 0 to 1. Throughout this paper, we use the term \emph{attribute} to refer to JSON keys that represent fields of the structured product data model.

The similarity function iterates over all attributes of the extracted object, comparing them with the corresponding values in the ground-truth instance provided by the data set. For primitive data types, equality is checked using the standard equality operator. For attributes associated with nested \texttt{NutritionMeasurementType} fields, the similarity function is applied recursively to both the extracted and ground-truth instances.
If the values of an attribute match, a local similarity score of $1$ is assigned. Otherwise, one of three error types is recorded: \textbf{AdditionalAttributeError}, \textbf{MissingAttributeError}, or \textbf{ValueError}, depending on the nature of the mismatch.

Each attribute is thus assigned a local similarity score between 0 and 1, depending on whether the extracted value matches the ground-truth value exactly or only partially (e.g., based on string similarity). The overall similarity score is then calculated as the average of all top-level attribute scores. This score reflects how closely the extracted instance matches the ground truth and serves as the basis for evaluating the performance of the extraction strategies.

\begin{table}[h]
    \centering
    \begin{tabular}{|c|c|c|}
        \hline
        \small{\textbf{Shop}} & \small{\textbf{HTML\_COMPRESSED}} & \small{\textbf{TEXT}} \\ \hline
        Globus & 97.77 & 96.00 \\ \hline
        Edeka24 & 98.46 & 96.69 \\ \hline
        \small{Supermarkt24h} & 98.05 & 95.44 \\ \hline
    \end{tabular}
    \caption{Accuracy values for the direct extraction conducted using the o3-mini model. The values have been calculated according to our similarity function.}
    \label{tab:precision-direct-extraction}
\end{table}

\subsection{\uppercase{Direct extraction}}
To evaluate the direct approach, we extracted information structured according to the response model from both compressed and plain text versions of the source files, as described in Section~\ref{sec:direct-extraction}. The results, presented in Table~\ref{tab:precision-direct-extraction}, indicate that the direct approach performs well for both document types, with slightly better performance on the \texttt{HTML\_COMPRESSED} version.

Due to the substantial costs involved, this extraction was performed only once, as depicted in Figure~\ref{fig:cost-comparison-direct-extraction}. The costs were calculated according to the pricing model outlined in Table~\ref{tab:price-table}.

\begin{table}
    \centering
    \begin{tabular}{|l|r|r|}
    \hline
    \textbf{Category} & \textbf{o3-mini} & \textbf{gpt-4o} \\ \hline
    Input (\$ / 1M Tokens) & 1.10 & 2.50 \\ \hline
    Cached Input (\$ / 1M Tokens) & 0.55 & 1.25 \\ \hline
    Output (\$ / 1M Tokens) & 4.40 & 10.00 \\ \hline
    \end{tabular}
    \caption{OpenAI API pricing specification\protect\footnotemark.}
    \label{tab:price-table}
\end{table}

\footnotetext{OpenAI Pricing: \url{https://openai.com/api/pricing/}, last visited 26.03.2025.}

\renewcommand{\thefigure}{3}
\begin{figure}
    \centering
    \includegraphics[width=1\linewidth]{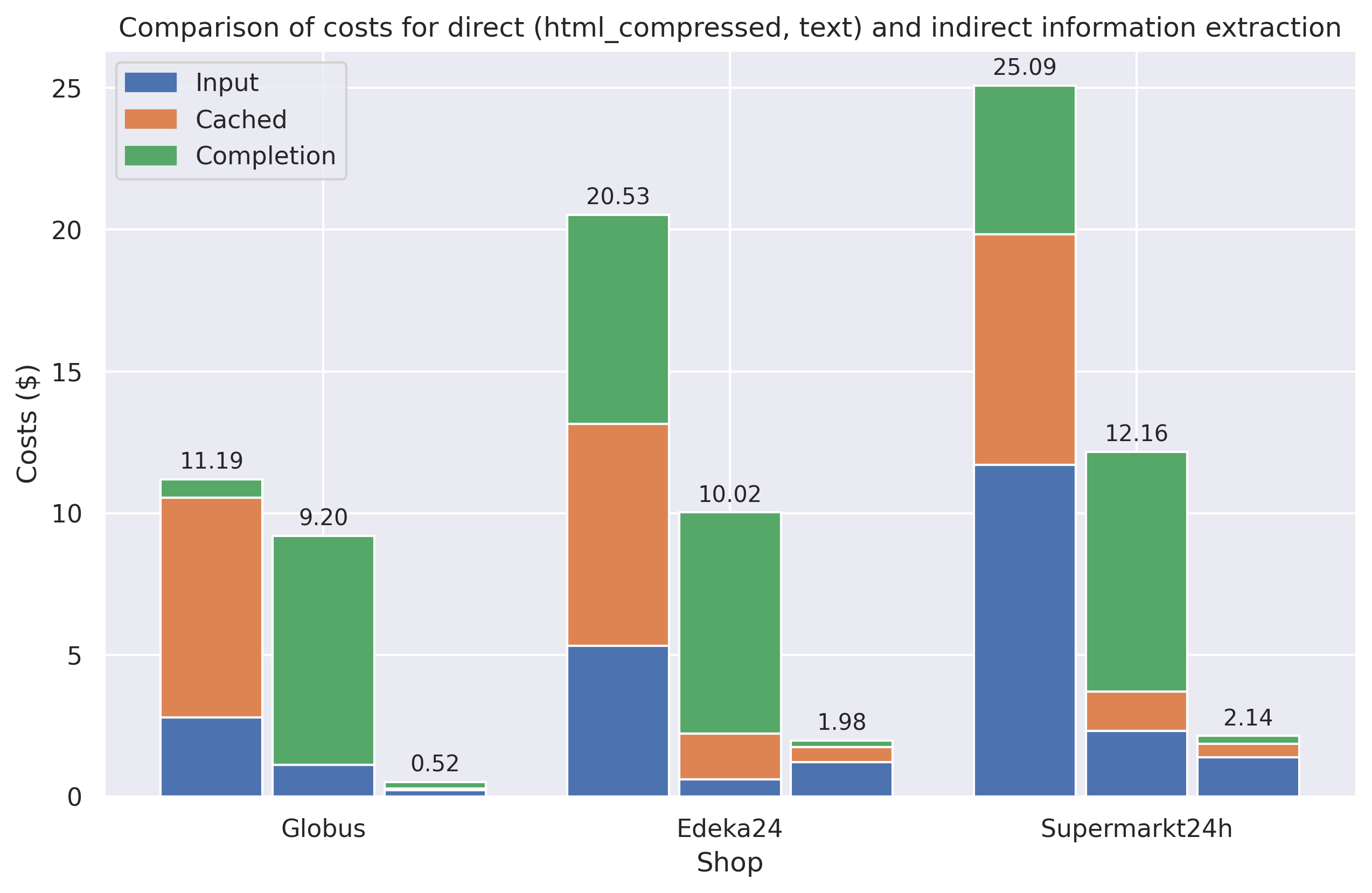}
    \caption{Costs occurred during the direct extraction for HTML\_COMPRESSED (left) and TEXT (middle) and the indirect extraction performed for HTML\_COMPRESSED (right). The indirect extraction has been performed ten times, costs are averaged.}
    \label{fig:cost-comparison-direct-extraction}
\end{figure}

\subsection{\uppercase{Indirect extraction}}
In our experiment, we executed the algorithm described in Section~\ref{sec:indirect-extraction} ten times for each shop, as we observed noticeable variability in the resulting accuracy scores. Figure~\ref{fig:indirect-approach-precision} shows these results and illustrates that all three shops can be successfully processed using the proposed algorithm; however, the accuracy values vary between runs and lack stability, reflecting the inherent variability of the approach.
This variability is primarily caused by the non-deterministic behavior of the \texttt{o3-mini} model during the generation and refinement of extraction functions. Additionally, randomizing the order of the products before each run may lead to the selection of suboptimal pages for initial function generation, further contributing to performance fluctuations.

To evaluate the quality of the decision functions generated by \texttt{gpt-4o}, we conducted an additional experiment for each shop, measuring the the classification accuracy against our data set. Across all shops, the decision functions achieved an average accuracy of 97.34\% over ten runs.

During each run, an average of 2.26 functions were generated and refined 3.96 times across all shops. Additionally, 36.06 direct extraction requests were made to find suitable reference objects for the function generation and refinement process. In total, this resulted in 44.18 calls to the \texttt{o3-mini} model per run. Furthermore, five additional requests were sent to the \texttt{gpt-4o} model to generate the decision function.

The average costs for each run are shown in the rightmost bars of Figure \ref{fig:cost-comparison-direct-extraction}. 

\renewcommand{\thefigure}{4}
\begin{figure}
    \centering
    \includegraphics[width=1\linewidth]{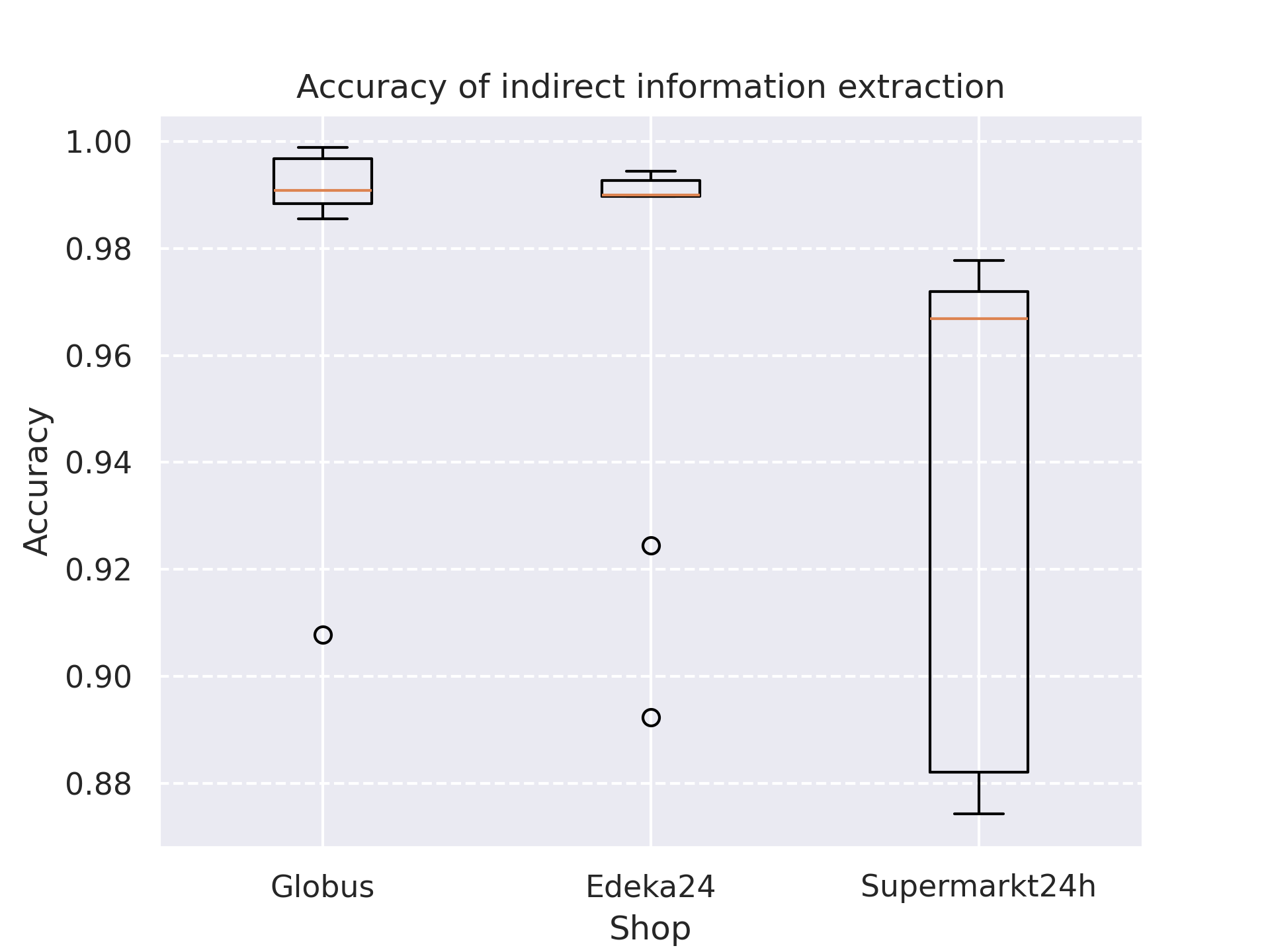}
    \caption{Accuracy value distribution for each shop across ten independent runs.}
    \label{fig:indirect-approach-precision}
\end{figure}

\newpage
\section{\uppercase{Discussion}}
\label{Discussion}

In this paper, we explored two approaches for extracting information from product web pages using large language models (LLMs): direct and indirect extraction. Both approaches adapt to user-provided data models and demonstrated strong performance in extracting nutritional information and ingredient lists.

The indirect approach offers significant efficiency gains by dynamically generating and refining extraction functions. Its performance is comparable to that of the direct approach, particularly for web pages following consistent templates, such as those used by online shops, internet forums, or news outlets. This consistency enables efficient iterative generation of reusable extraction functions and substantial cost reductions in large-scale web scraping scenarios. For example, while the direct approach required 1,000 requests per run for a single shop, the indirect approach averaged only 44.18 requests, a reduction of $95.82\%$ when querying \texttt{o3-mini}, not including the five static requests to \texttt{gpt-4o} for the decision function.

While direct extraction generally incurs higher costs, it achieves slightly better overall accuracy ($+1.61\%$). One notable observation from the direct extraction experiments was a slight decline in accuracy for one shop, attributed to missing ingredient statements for certain products such as fresh fruits and vegetables. In these cases, the model (\texttt{o3-mini}) tended to hallucinate attributes, for instance by inferring ingredients from simple product names (e.g., "Braeburn apple").

Variability in accuracy was also observed in indirect extraction experiments, indicating room for further improvement. Refining the attribute-level or aggregated meta-attribute decision procedures could enhance performance.

\renewcommand{\thefigure}{5}
\begin{figure}[h]
    \centering
    \includegraphics[width=1\linewidth]{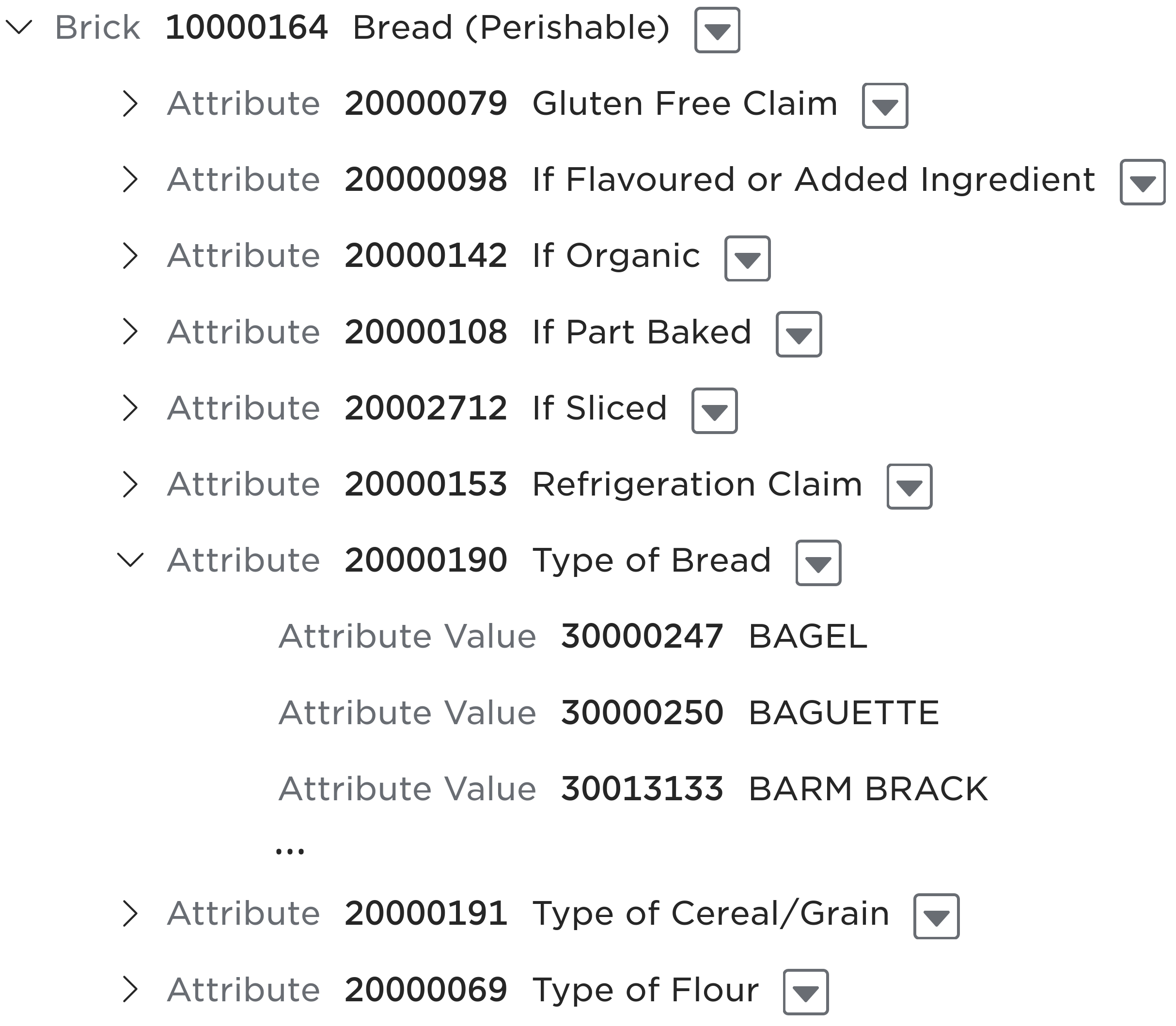}
    \caption{Attribute list for the GPC brick code 10000164 - Bread (Perishable), showing a subset of defined attribute values for attribute 20000190 - Type of Bread.}
    \label{fig:gpc-attribute-list}
\end{figure}

For template-based product pages, the indirect approach appears to be the more scalable option, as its efficiency depends on the number of unique templates rather than the number of individual pages.
However, broader applicability remains to be tested. Certain product attributes, such as those defined by the GPC taxonomy, may not always be accessible via CSS or XPath selectors, as they can be nested within product descriptions or other unstructured text rather than explicitly encoded in the HTML source. 

As illustrated in Figure~\ref{fig:gpc-attribute-list}, the GPC attribute list for bread products contains detailed attribute values that are often embedded in unstructured formats. This complexity suggests that, for such cases, direct extraction may outperform indirect extraction approaches.

\section{\uppercase{Conclusion}}
\label{Conclusion}

Our comparative study shows that both direct and indirect LLM-based extraction approaches can effectively automate information retrieval from product web pages. The indirect approach offers substantial cost savings and comparable accuracy, particularly when pages follow largely consistent templates, even if minor structural variations exist within a single shop.

Future work will focus on evaluating a wider range of LLMs, refining the dynamic function generation process, and expanding applicability to more diverse web structures and attribute types. In particular, our goal is to utilize attributes defined in the GPC taxonomy with our methods by reliably classifying products to their lowest hierarchy level (brick), thereby improving automated attribute extraction from complex product pages.

Ultimately, our results suggest that LLM-based automation could serve as a practical and scalable alternative to manually defined web scraping, enabling seamless integration with existing data models and a wide range of downstream applications.


\section*{\uppercase{Code Availability}}
All Jupyter notebooks, scripts, and data can be found in repositories within the following group: \url{https://gitlab.rlp.net/ISS/smartcrawl}.

\section*{\uppercase{Acknowledgements}}

This work was funded by the German Federal Ministry of Education and Research, BMBF, FKZ 01\textbar S23060.

Parts of the text have been enhanced and linguistically revised using artificial intelligence tools. All concepts
and implementations described are the intellectual work of the authors.

\bibliographystyle{apalike}
{\small
\bibliography{example}}
\end{document}